*published at NCTA 2024, Porto, Portugal*

# Contrastive Learning and Abstract Concepts: The Case of Natural Numbers

Daniel N. Nissani (Nissensohn)[1]
[1]*Independent Research*
*dnissani@post.bgu.ac.il*

Abstract: Contrastive Learning (CL) has been successfully applied to classification and other downstream tasks related to concrete concepts, such as objects contained in the ImageNet dataset. No attempts seem to have been made so far in applying this promising scheme to more abstract entities. A prominent example of these could be the concept of (discrete) Quantity. CL can be frequently interpreted as a self-supervised scheme guided by some profound and ubiquitous conservation principle (e.g. conservation of identity in object classification tasks). In this introductory work we apply a suitable conservation principle to the semi-abstract concept of natural numbers by which discrete quantities can be estimated or predicted. We experimentally show, by means of a toy problem, that contrastive learning can be trained to count at a glance with high accuracy both at human as well as at super-human ranges.. We compare this with the results of a trained-to-count at a glance supervised learning (SL) neural network scheme of similar architecture. We show that both schemes exhibit similar good performance on baseline experiments, where the distributions of the training and testing stages are equal. Importantly, we demonstrate that in some generalization scenarios, where training and testing distributions differ, CL boasts more robust and much better error performance.

## 1 INTRODUCTION AND RELATED WORK

Contrastive Learning (CL) is a self supervised scheme which has attracted much attention in recent years. In the visual modality realm it maps a visual input (e.g. objects to be classified) to linearly separable representations which achieve classification accuracy rates competitive with those of supervised learning (SL) networks of similar architecture (Chen et al., 2020) in challenging datasets such as ImageNet (Deng et al., 2009).

In spite of its impressive success in the space of concrete concepts (Chen et al., 2020), neither CL nor its variants (Grill et al., 2020; He et al., 2020; Chen & He, 2020) have been apparently applied so far to the learning and prediction of abstract or semi-abstract entities. In a recent work (Nissani (Nissensohn), 2023) has shown that CL can (unlike SL) build "hyper-separable" representations which are useful not only to predict an object identity but also to indicate the existence (or absence) of selected attributes of interest of this object; this might be seen as a first modest step away from the concrete and towards the abstract. Another prominent example of the learning, 'grounding', or in-depth 'understanding' of such an abstract entity could be that of the concept of natural numbers (equivalently, discrete quantities). This work is a preliminary and introductory step forward in this direction.

CL exploits, in the concrete visual modality, a profound principle of conservation: that distinct views of an object preserve the identity of said object. To create such distinct views suitable transformations should (and generally can) be designed (Tian et al., 2020). Analog principles of conservation have been applied in physics (e.g. conservation of energy, of momentum, etc. under suitable reference frames transformations) with extraordinary success during the last two centuries.

To apply a relevant and useful transformation to our (discrete) quantity prediction challenge within a CL scheme we can imagine the following thought experiment: that we have objects which we wish to count at a glance (that is "count without counting", see ahead), that these objects lay at the bottom of a closed box with transparent cover. Shaking the box, our transformation, will randomly change the layout of the objects inside the box, but the total number of objects on the box floor will be conserved (since the box is closed, etc.).

Our CL optimization goal *l(i, j)* will then consist of minimizing the normalized distance (Wang & Isola, 2020) between the neural network so called projection (i.e. last) layer (Chen et al., 2020) representations $z_i$ and $z_j$ of the pre- and post-shaking views ('positive samples' in the CL jargon) while simultaneously maximizing the distances between these representations and the representations of other samples randomly gathered in a mini-batch ('negative samples'). Formally, our goal will then be (see Chen et al., 2020 for details):

$$l(i,j) = -\frac{sim(z_i, z_j)}{\tau} + \log\left(\sum_{k=1}^{2N} 1_{k\neq i} \exp\left(\frac{sim(z_i, z_k)}{\tau}\right)\right) \quad (1)$$

$$L = \frac{1}{2N}\sum_{k=1}^{N}[l(2k-1, 2k) + l(2k, 2k-1)] \quad (2)$$

where *sim*(. , .) is the normalized inner product (i.e. the cosine similarity), $\tau$ is a system temperature, N is the mini-batch size, $1_{k\neq i}$ is a binary indicator function which vanishes when k = i and equals 1 otherwise, and *L* is the overall loss, i.e. *l(i, j)* summed over all samples in the mini-batch. At the end of the optimization process we freeze the neural network learned parameters, fetch an interior layer and define its output to be our linearly separable representation vectors. These are then fed into a (usually supervised, single layer) linear classifier.

We humans are able to estimate at a glance, with high precision and with no explicit enumeration, a relatively small (up to between 4 to 7) number of objects in view (Trick & Pylyshyn, 1994). This capability, for which the special term 'subitizing' was coined (Kaufman et al., 1949) has motivated in recent years a few groups of neural networks practitioners (Chattopadhyay et al, 2017; Acharya et al., 2018) to explore the application of supervised learning (SL) schemes to a similar challenge.

We are not aware of any similar work on natural numbers under the umbrella of CL, nor of the application of CL schemes to other non-tangible concepts.

As a preliminary introduction to these ideas we implement a toy problem and corresponding datasets by means of which CL and SL networks of similar architecture are trained to subitize and predict the quantity of identical objects present in an image. We compare the error performance of these two schemes at both a 'baseline' regime, where training and testing data originate from identical distributions, as well as at various generalization regimes, where training and testing data distributions differ. CL is trained by a more profound, potentially 'grounding', criterion than SL, a criterion that is intimately related to the concept of Quantity itself. We thus may well suspect that it will exhibit error performance better than that of SL, at least in some of the forementioned generalization regimes.

The main contributions of this introductory work are:

a. We demonstrate through the example of the Quantity (equivalently, Natural Numbers) non-tangible concept, that CL can learn abstract concepts whenever transformations which conserve their defining properties can be identified and implemented.
b. We show that both CL and SL can learn to subitize, both at human range (order of 10 objects) and super-human range (order of 100). Moreover, that the error performance of CL and SL are similarly good under the forementioned baseline regime scenarios.
c. We show that CL exhibits significant error performance superiority over SL under certain generalization regime scenarios, possibly corroborating, as suspected, the more profound and grounded nature of learning by means of a conservation principle (relative to SL learning which consists of merely forcing representations to adjust to arbitrary labels).

In Section 2 the toy problem and test setup which we introductorily employ to demonstrate these ideas are presented. Section 3 details our simulation results, and Section 4 provides concluding remarks and outlines potential future lines of research.

## 2 EXPERIMENTAL SETUP

We implement a toy problem to conduct experiments, demonstrate our ideas and probe into our CL superiority conjecture. We generate (practically infinite) sequences of random synthetic images of dimension d x d pixels, with d = 22 or 28. Each image contains a number of identical white objects laid out over a black background. For each experiment we select a training dataset distribution

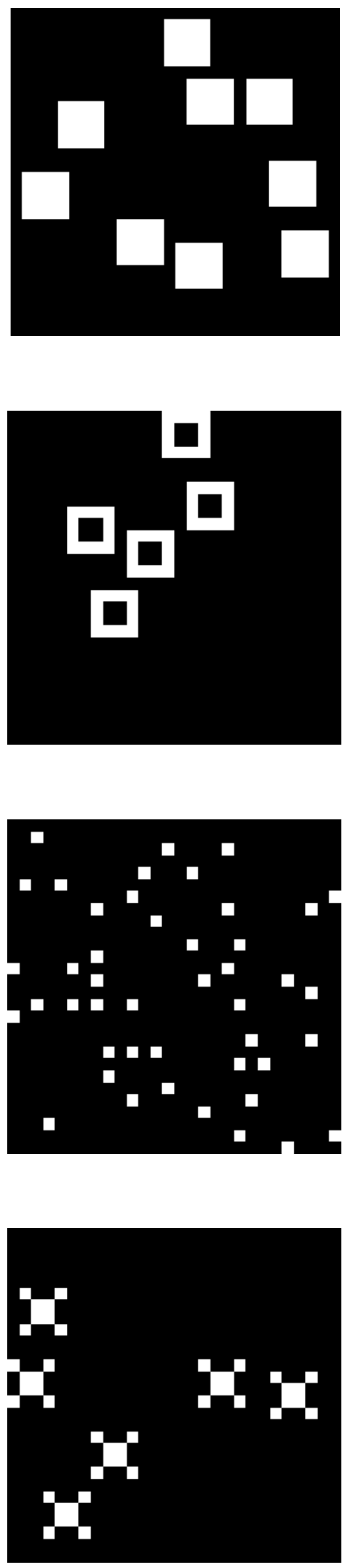

Figure 1. Synthetic image samples. Each image of d x d pixels (d = 28 in this Figure) contains a random layout of a random number of objects of an identical shape. Shapes are either (from top to bottom) Full squares (denoted F in the text), Void squares (V), Dots (D) and Crosses (C); we may either preset the objects shape for all images within the dataset with the same selected shape, or randomly pick for each image a fixed shape amongst F/V/C which we denote by X (for miXed, not shown in Figure). The shapes definition is thus an element of {F, V, C, D, X}.

{O, S, R} and a testing dataset distribution {O', S', R'} so we designate an experiment by the composite triplet {O/O', S/S', R/R'}.

Baseline experiments consist of triplets where O = O', S = S' and R = R'; otherwise they are Generalization experiments wherein the training and testing distributions differ.

O is the maximal number of objects in an image and took in our experiments the values 10 (to emulate human range), 60 or 80 (for super-human); the number of objects in an image is uniformly randomly selected within a subset of [1, O].

S defines the shape of all objects within an image and, for the sake of some shape diversity, can take the values F, V, C, D or X (see Figure 1 for details).

R can take the values A (for All), E (for Evens) or O (for Odds) where A means that the number of objects in an image can be any in the range [1, O] while E or O restricts this number to the subset of even or odd values respectively.

A typical experiment may consist of {O/O', S/S', R/R'} = {80/80, D/D, A/A} which means identical training and testing distribution (i.e. a Baseline experiment) with between 1 to 80 objects in each image, each object of Dot shape, and no Evens nor Odds only restriction.

The objects are randomly laid out within the image. We do not allow objects to occlude nor touch each other. Each sample image is generated along with its label which describes the number of objects in that sample.

We train a simple fully connected multi layer neural network by CL with architecture [$d^2$ 400 400 100] where the last layer, of dimension 100, is the (so called) projection layer (Chen et al., 2020) and the penultimate layer, of dimension 400 is our representation layer.

We use for SL an identical architecture, except for the last layer which for SL contains a softmax activation function of dimension O. Note that we could instead have chosen to implement a regression (e.g. linear) scheme for SL. We opted for the former since it supports a more 'apples to apples' comparative study.

It is not so simple to physically emulate the "shaking of our box" (of our thought experiment above) in order to create a new image sample with the same number of objects within it but with a different random objects layout. Instead, we opt for a surrogate: we take the label attached to each image (which describes the number of objects in that image) and use it to generate another (random layout) image with this specified number of objects.

This may seem at first glance a cheating perversion by which we convert our CL (unsupervised learning) scheme into a supervised learning one (since we are now using labels for our ‘transformation’). After some reflection however, it should be easy to conclude that this is immaterial to our purpose. The same results we are going to show can be exactly replicated by a true physical shaking of our box.

Training of CL or SL proceeded for a number of samples until no further visible convergence progress is observed in the Loss goal (Eqns. (1) and (2) above for CL) or in the training Mean Squared Error goal (for SL). Once training is halted we freeze the CL (or SL) parameters, and train a single layer supervised linear classifier (Oord et al., 2018) with the CL (or SL) network representations generated by the testing dataset distribution. After this linear classifier is trained, we evaluate error performance with samples generated by means of this same selected test dataset distribution (as mentioned above our datasets are practically infinite so we are not, by no means and as properly prohibited, re-using samples for both training and testing).

We set the CL hidden layers neural units activation to ReLU and the last layer units to tanh. CL temperature ($\tau$ of Equation 1) was set to 1, and batch size (N of Equation 2) was set to $10^3$. We used ADAM (Kingma & Ba, 2015) gradient descent with fixed learning rate $\eta = 10^{-3}$. The number of training samples ranged from $1.8 \times 10^6$ (equivalent to e.g. 30 MNIST epochs) to $3.6 \times 10^6$ depending on the running experiment.

SL units were here again set to ReLU for hidden layers but to softmax activation for the last layer. We used ADAM with fixed learning rate $\eta = 5 \times 10^{-4}$ for the initial stage followed by plain gradient descent with learning rate typically descending from $10^{-5}$ to $10^{-7}$. We found this protocol necessary to achieve our best error performance, possibly testifying the navigation within a deep narrow valley (Martens, 2010). The number of SL training samples ranged from $4.2 \times 10^6$ to $10.8 \times 10^6$.

Finally, the linear classifier was trained by ADAM followed by plain SGD, with similar optimization protocol as that of SL above and number of samples typically ranging from $2.4 \times 10^6$ to $28.8 \times 10^6$.

Across all three above schemes ADAM parameters were set to $\beta_1 = 0.9$, $\beta_2 = 0.999$ and $e = 10^{-8}$.

To facilitate replication of our results a simulation package will be provided by the author upon request.

# 3 SUBITIZING EXPERIMENTAL RESULTS

In this Section we will be comparing CL vs. SL subitizing error performance. To probe whether our forementioned conjecture, which states that CL will boast superior performance w.r.t. SL in at least some of the Generalization scenarios (where training and test distributions differ) we define a set of appropriate experiments. We also provide baseline results (where training and testing distributions are identical) for reference.

Generalization here can be applied along the shape dimension, or the quantity dimension, or both at a time (which we do not pursue herein). For convenience we list below the experiments (generalized item bold and underlined; please refer to Section 2 for triplets legend):

- Shape Dimension (human range; train with miXed images, test with Dots):
  - Baseline: {O/O’, S/S’, R/R’} = {10/10, X/X, A/A}
  - Generalization: {O/O’, S/S’, R/R’} = {10/10, <u>**X/D**</u>, A/A}
- Quantity Dimension, Range extension (super human range; train with up to 60 objects, test with up to 80):
  - Baseline: {O/O’, S/S’, R/R’} = {60/60, D/D, A/A}
  - Generalization: {O/O’, S/S’, R/R’} = {<u>**60/80**</u>, D/D, A/A}
- Quantity Dimension, within Range (super human range; train with up to 80 objects, Even values only, test with up to 80, All values):
  - Baseline: {O/O’, S/S’, R/R’} = {80/80, D/D, E/E}
  - Generalization: {O/O’, S/S’, R/R’} = {80/80, D/D, <u>**E/A**</u>}
- Quantity Dimension, within Range (super human range; train with up to 80 objects, Even values only, test with up to 80, Odd values only):
  - Baseline: {O/O’, S/S’, R/R’} = {80/80, D/D, E/E}
  - Generalization: {O/O’, S/S’, R/R’} = {80/80, D/D, <u>**E/O**</u>}

### 3.1 Shape Dimension (human range; train with miXed images, test with Dots)

We report $Pr^{CL, B}$\{error\} = 0.042 and $Pr^{SL, B}$\{error\} = 0.007 for Baseline test (denoted by B in super-script), i.e. {O/O', S/S', R/R'} = {10/10, X/X, A/A}, for CL and SL respectively.

That SL yields better performance than CL under baseline tests with similar networks architecture is no surprise: it was noticed in prior works, see e.g. (Chen et al. 2020) where in order to achieve similar error performance a deeper and wider architecture (on CL network relative to SL) was required.

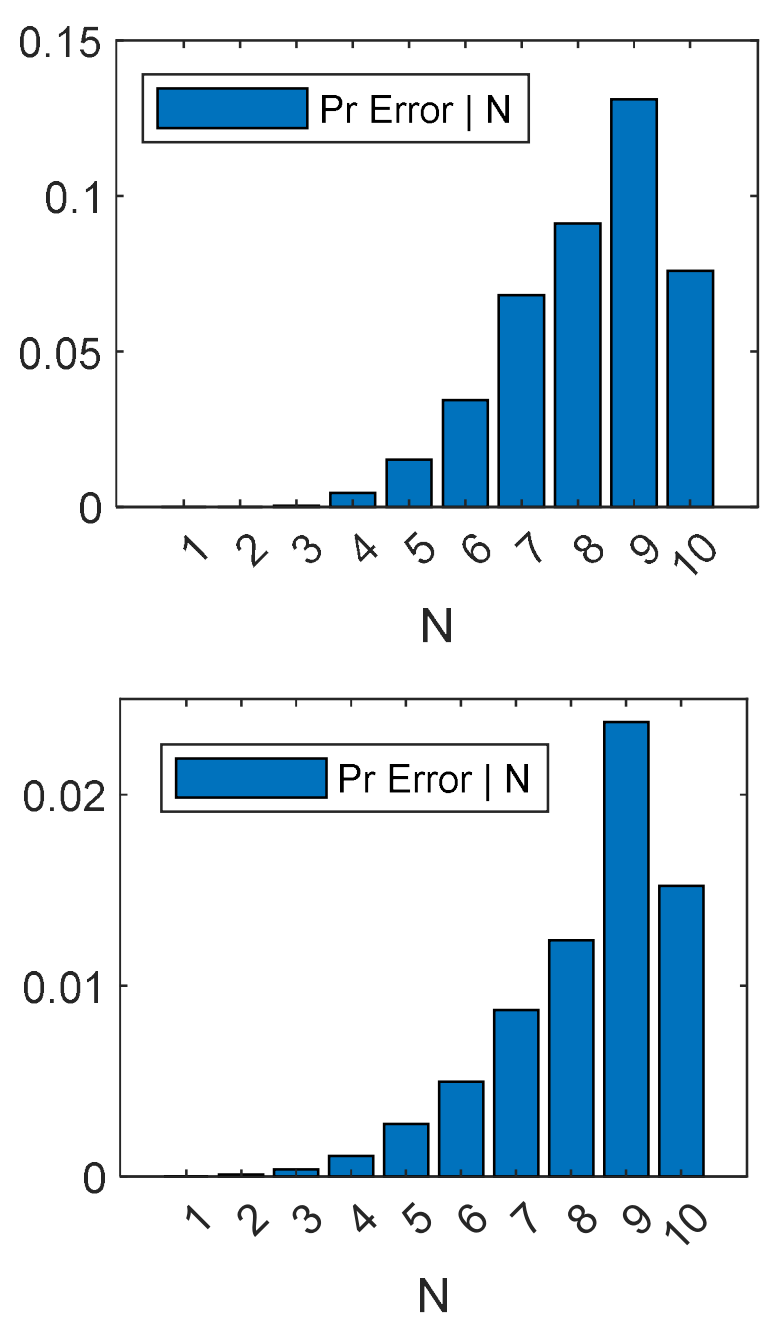

Figure 2. Shape Baseline test, 10 Objects, train with miXed, test with miXed. Conditional probability of Error given Ground Truth for CL (top) and SL (bottom).

Interestingly, errors are not uniformly distributed across ground truth values at both CL and SL (see Figure 2): as might be intuitively expected error rate steadily increases from small to large ground truth values (with 10 a possible curious exception; this may be an artifact of our setup: since we set a hard limit of 10 through our softmax function, errors at this ground truth value are contributed, unlike at other values, by one side only). A qualitatively similar phenomenon was observed in subitizing experiments conducted with human subjects (Kaufman et al., 1949). We did not observe however a similar trend with our experiments at super human range.

We next turn to the corresponding Shape Generalization test, i.e. {O/O', S/S', R/R'} = {10/10, **X/D**, A/A} where training is conducted with miXed images (which, as described above, contain either F or V or C objects but do not contain Dots) but testing is done with Dots images. Error probability significantly degrades, with similar degradation at CL and SL: $Pr^{CL, G}$\{error\} = 0.31 and $Pr^{SL, G}$\{error\} = 0.30 (superscript G here denotes Generalization).

More important perhaps however in a quantity estimation task than this gross error measure is the conditional distribution of Distances (conditioned on error events), where Distance = |Ground Truth value – Predicted value|: in practical situations to predict a value of 8 instead of a ground truth value 9 is forgivable while to predict a 1 in the same case is not.

This more suitable metric can be observed in Figure 3: we notice that in spite the significant forementioned Pr\{error\} degradation, the distribution of Distance in both CL and SL is remarkably concentrated in the very low values, with $Pr^{CL\ G}$\{Distance > 1 | error\} = 0.036 and $Pr^{SL\ G}$\{Distance > 1 | error\} = 0.041, again pretty similar to each other.

To summarize, in this Shape Generalization experiment CL and SL perform similarly; they both degrade significantly in terms of Pr\{error\}, but their more tolerant (and perhaps relevant) Pr\{Distance > 1 | error\} metric is still pretty good.

We have provided this Figure to contrast vs. next experiments, as we will immediately see.

### 3.2 Quantity Dimension, Range Extension (super human range; train up to 60 objects, test with up to 80)

Here we report $Pr^{CL, B}$\{error\} = 0.029 and $Pr^{SL, B}$\{error\} = 0.004 for the Baseline test, i.e. {O/O', S/S', R/R'} = {60/60, D/D, A/A}, for CL and SL respectively. Since the Distance distribution metrics is apparently of more significant relevance in subitizing tasks we also report for the Baseline test $Pr^{CL\ B}$\{Distance > 1 | error\} = 0.035 and $Pr^{SL\ B}$\{Distance > 1 | error\} = 0.002.

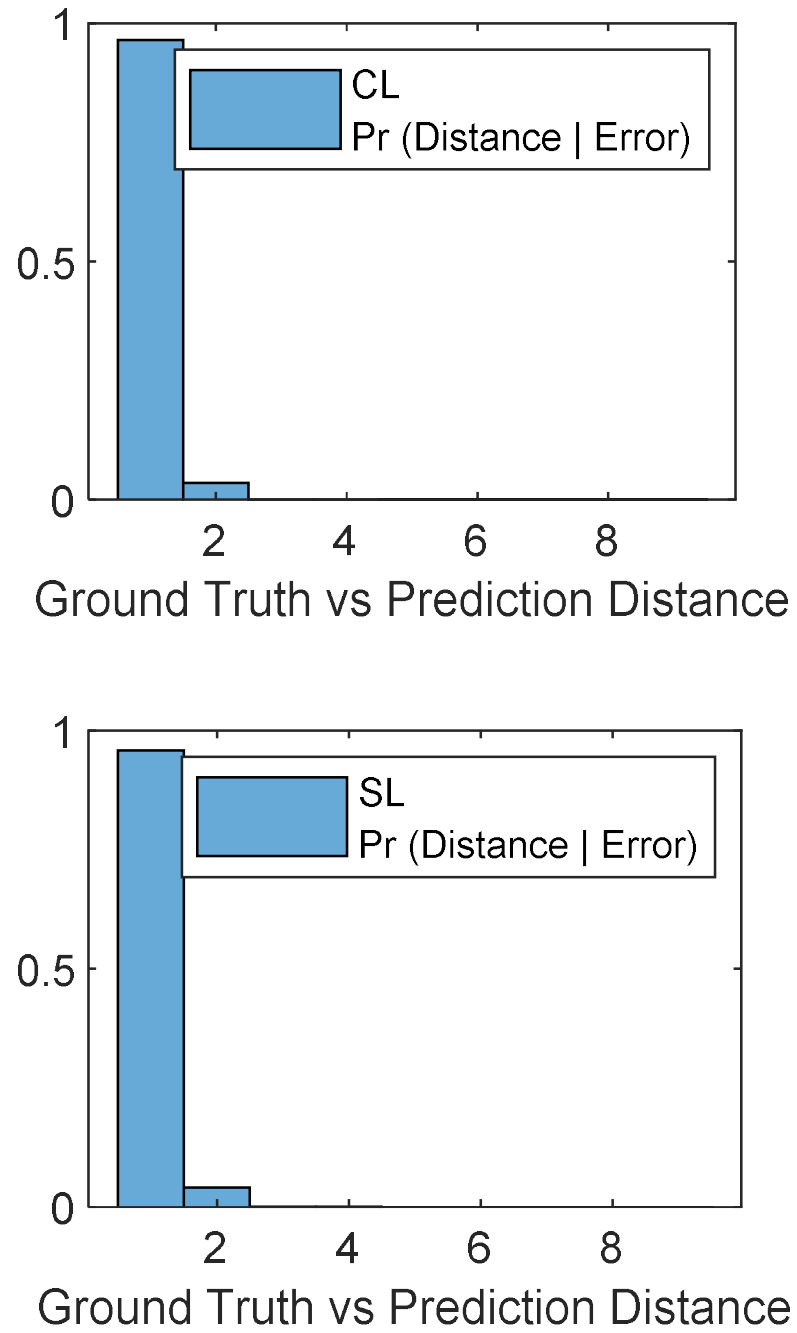

Figure 3. Shape Generalization, train with miXed, test with Dots. Conditional probability distribution of Ground Truth vs. Predicted Distance for CL (top) and SL (bottom).

The corresponding Quantity Generalization test is {O/O', S/S', R/R'} = {**<u>60/80</u>**, D/D, A/A} where training is conducted with up to 60 Dots images and testing is done with up to 80 Dots images. Again, as in the Shape Generalization experiment, error probability significantly degrades, with similar degradation at CL and SL: $Pr^{CL, G}$\{error\} = 0.24 and $Pr^{SL, G}$\{error\} = 0.23.

Conditional Distance distributions of CL and SL, however, are totally different. Please refer to Figure 4. While CL Distance distribution is concentrated in the very low values (as in our earlier experiments) and $Pr^{CL\ G}$\{Distance > 1 | error\} = 0.12, SL distribution practically explodes with $Pr^{SL\ G}$\{Distance > 1 | error\} = 0.80.

Briefly summarizing so far, in shape generalization experiments (Subsection 3.1 above) both CL and SL behave similarly: this is reasonable since it is not shape what CL learns in depth and thus they should not differ much from each other, just as in baseline tests.

In contrast, in quantity generalization scenarios, CL and SL respond extremely differently and CL significantly outperforms SL. This can be attributed in our view to the profoundness to which CL learns the abstract concept of "numberness" by means of the contrastive learning conservation principle, compared to the merely forced mapping of input images to arbitrary labels in SL. The concept of Numbers seems to become 'grounded' in CL but not so in SL.

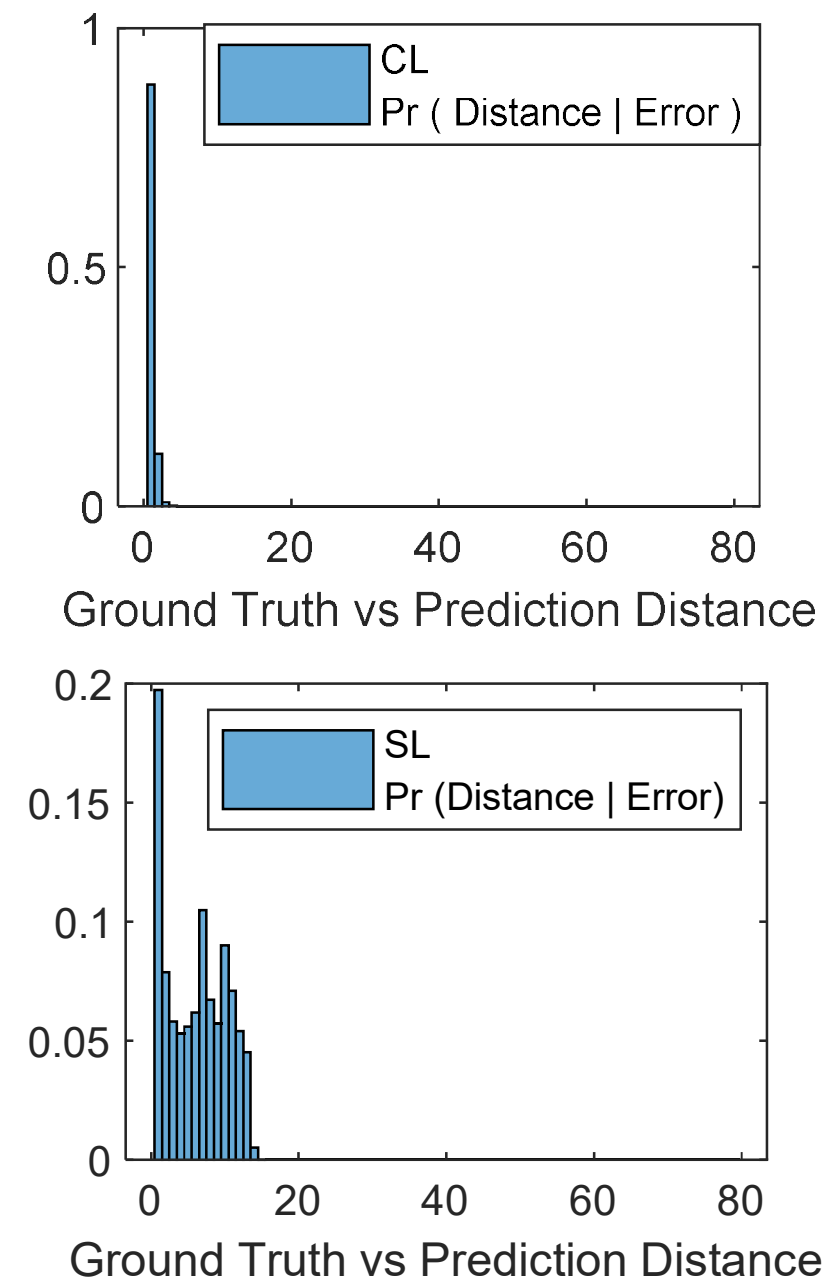

Figure 4. Quantity Generalization, train with up to 60, test with up to 80 Dots images. Conditional Distance distribution for CL (top) and SL (bottom). CL exhibits a modest degradation w.r.t. baseline test while SL collapses.

### 3.3 Quantity Dimension, within Range (super human range; train with up to 80 objects, Even values only, test with up to 80, All values)

Next we report $Pr^{CL, B}$\{error\} = 0.019 and $Pr^{SL, B}$\{error\} = 0.0004 as well as $Pr^{CL\ B}$\{Distance > 2 | error\} = $Pr^{SL\ B}$\{Distance > 2 | error\} = 0 for the Baseline test of this experiment, i.e. {O/O', S/S', R/R'} = {80/80, D/D, E/E}, for CL and SL respectively. Notice that conditional Distance distribution results here are reported w.r.t. a threshold valued 2 (rather than 1 as before) since Odds "do not exist" in this Baseline scenario.

The corresponding Quantity Generalization test is {O/O', S/S', R/R'} = {80/80, D/D, **<u>E/A</u>**} where training is conducted with even quantities (i.e. 2, 4.

6…80) of up to 80 Dots images, and testing is done with all quantities (i.e. 1, 2, 3….80) of up to, again, 80 Dots images. SL degrades its error probability w.r.t. baseline here, while CL this time exhibits even better performance than its own baseline: $Pr^{SL\ G}$\{error\} = 0.045 and $Pr^{CL\ G}$\{error\} = 0.00003.

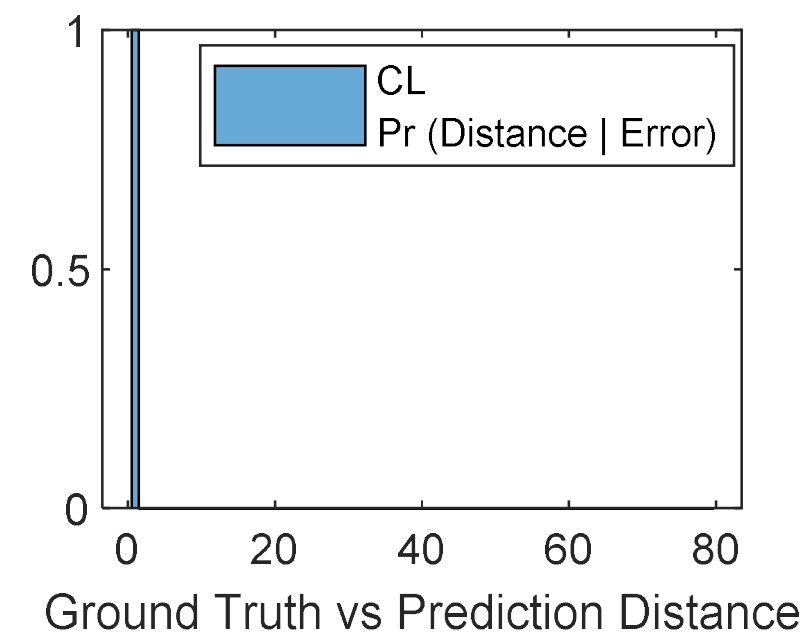

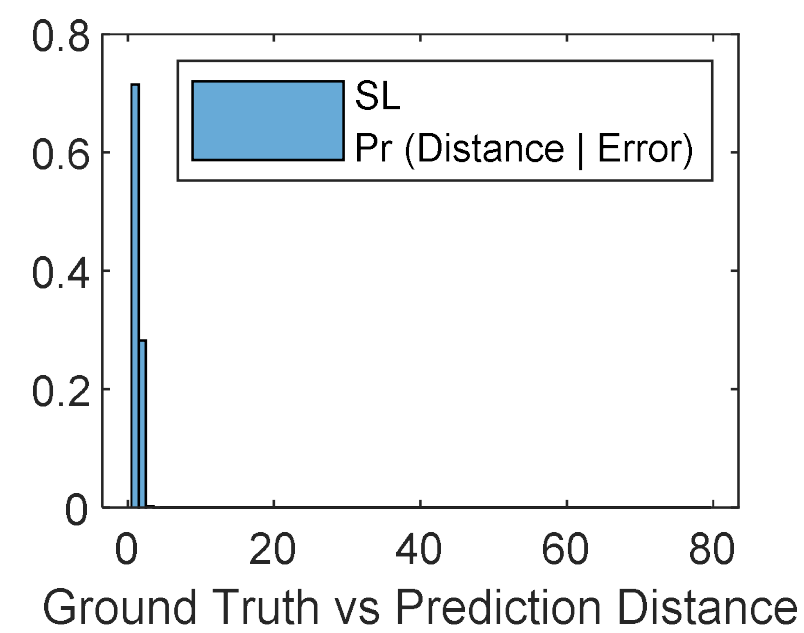

Figure 5. Quantity Generalization, train with Evens only, test with All, both up to 80 Dots images. Distance distribution for CL (top) and SL (bottom). CL exhibits no degradation w.r.t. baseline test while SL once again collapses.

Evaluating again with our conditional Distance distribution metrics (please refer to Figure 5), CL significantly outperforms SL here too with $Pr^{CL\ G}$\{Distance > 1 | error\} = 0 and $Pr^{SL\ G}$\{Distance > 1 | error\} = 0.285, supporting once again our motivating conjecture.

### 3.4 Quantity Dimension, within Range (super human range; train with up to 80 objects, Even values only, test with up to 80, Odd values only)

Our baseline setup here is identical to that of the last experiment, i.e. {O/O', S/S', R/R'} = {80/80, D/D, E/E}, for CL and SL respectively, and so are our results. The conditional Distance distributions results here (for both baseline and generalization tests) are once again reported w.r.t. a threshold valued 2 (rather than 1) for the same reason as in Subsection 3.3 above.

The corresponding Quantity Generalization test is {O/O', S/S', R/R'} = {80/80, D/D, **<u>E/O</u>**} where training is conducted with even quantities (i.e. 2, 4. 6…) of up to 80 Dots images and testing is done with odd quantities (i.e. 1, 3, 5…79) of up to 79 Dots images. Both CL and SL exhibit good error probability with $Pr^{CL\ G}$\{error\} = 0.026 and $Pr^{SL\ G}$\{error\} = 0.026 and both show excellent Distance conditionals with $Pr^{CL\ G}$\{Distance > 2 | error\} = $Pr^{SL\ G}$\{Distance > 2 | error\} = 0.

It appears that once CL and SL learn the Evens the discriminative resolution of the resulting representations is good enough to predict the Odds with good error performance, For the SL scheme however, this holds only **<u>provided that</u>** the Evens are omitted from the testing dataset, so that they do not 'confuse' the SL network as apparently occurred in our previous Subsection 3.3 experiment.

Table 1 provides a concise summary of our results above.

## 4 DISCUSSION AND TOPICS FOR FUTURE RESEARCH

We have provided preliminary demonstrative evidence, through our selected example of quantity estimation at a glance, that contrastive learning methods can deal not only with concrete, tangible concepts. Our choice however of natural numbers is not casual. When dealing with concrete objects it is not difficult to identify transformation sets, with random properties, which efficiently span the distribution of a dataset; examples include crop and color distortion for ImageNet (Chen et al., 2020) and elastic distortion (Simard et al, 2003) for EMNIST (Nissani (Nissensohn), 2023). This is possibly not the case for abstract concepts in general, and identifying viable abstract sets and their corresponding spanning random transformations is a research challenge. It would be of interest to see

Table 1. Experiments results summary. T denotes a Threshold, valued 1 for all tests, except 5, 7 and 8 for which T = 2 (see text). Most informative results are **bolded/underlined**.

| Test # | EXPERIMENT | TEST | CL | SL | CL | SL |
|---|---|---|---|---|---|---|
| | | | **Pr{error}** | | **Pr{Distance > T \| error}** | |
| **1** | **Shape, train miXed test Dots** | **Baseline** | 0.042 | 0.007 | 0 | 0 |
| **2** | | **Generalized** | 0.31 | 0.30 | 0.036 | 0.041 |
| **3** | **Qty, Range extension, train 60 test 80 objects** | **Baseline** | 0.029 | 0.004 | 0.035 | 0.002 |
| **4** | | **Generalized** | 0.24 | 0.23 | **0.12** | **0.80** |
| **5** | **Qty, within Range, train Evens test All** | **Baseline** | 0.019 | 0.0004 | 0 | 0 |
| **6** | | **Generalized** | 0.00003 | 0.045 | **0** | **0.285** |
| **7** | **Qty, within Range, train Evens test Odds** | **Baseline** | 0.019 | 0.0004 | 0 | 0 |
| **8** | | **Generalized** | 0.026 | 0.026 | **0** | **0** |

other such concepts to follow-on our current very first step.

We find it quite remarkable that both CL and SL exhibit such impressive subitizing error performance in baseline test scenarios even at super-human ranges. We did not inquire into the existence of a practical upper bound to this range nor, more generally, into the error performance as function of the dataset range, which of course should be of interest.

The results shown indicate that CL and SL perform quite similar to each other in shape generalization tests, both with significant error probability degradation but with robust figures w.r.t. conditional distance distribution. This might be expected since the natural numbers conservation principle guiding CL should not grant it any advantage when dealing with varying shapes.

In contrast, CL alone maintains this robustness in quantity generalization tests as well, and in particular in range extension tests, where the scheme is asked to estimate at a glance a quantity bigger than what it was ever exposed to. This seems to support our motivating conjecture that stated that CL, because it is guided by the principle of conservation of natural numbers should obtain, after training, a more deep and grounded sense of ‘numberness’. And this would then be the second case in a row at which CL seems to show a fundamental superiority over SL in generating information rich representations (the first being the forementioned capability of detection of objects attributes (Nissani (Nissensohn), 2023)).

In this introductory work we have devised and employed a toy problem to demonstrate our ideas because it has better control of test bench variables and parameters. It should be of course important to evaluate whether similar results can be achieved in more realistic scenarios which may include clutter, occlusion, composite scenes with different classes of real-life objects, etc. Several datasets for such purpose are already available (Acharya et al., 2019).